%% file: root.tex
\documentclass[letterpaper, 10 pt, journal, twoside]{IEEEtran}

\usepackage{amsmath} 
\usepackage{amssymb}
\usepackage{amsthm}
\usepackage{bm}
\usepackage{mathrsfs}
\usepackage{graphicx}
\usepackage{epsfig}
\usepackage{subfigure}
\usepackage{enumerate}
\usepackage{cite}
\usepackage{setspace}
\usepackage{cancel} 
\usepackage{color}
\usepackage{wrapfig}
\usepackage{xspace}
\usepackage{paralist}
\usepackage[colorlinks, citecolor=black, linkcolor=black, linktocpage=true]{hyperref}
\usepackage{bookmark}
\usepackage[printonlyused]{acronym} 
\usepackage{parskip}
\usepackage{times}
\usepackage{array}
\usepackage{fixltx2e}
\usepackage{stfloats}
\usepackage{framed}
\usepackage{multirow}
\usepackage{accents}
\usepackage{float}
\usepackage{xspace}
\usepackage{siunitx}

\newcommand{\ignore}[1]{}

\newcommand{\norm}[1]{\left\Vert#1\right\Vert} 

\newcommand{\bma}[1]{\left[\begin{array}{#1}}
\newcommand{\ema}{\end{array}\right]}

\DeclareMathAlphabet{\mbf}{OT1}{ptm}{b}{n}
\newcommand{\mbs}[1]{{\boldsymbol{#1}}}

\newcommand{\mbsdot}[1]{{\dot{\boldsymbol{#1}}}}

\newcommand{\mbshat}[1]{{\hat{\boldsymbol{#1}}}}

\newcommand{\mbfdot}[1]{{\dot{\mbf{#1}}}}

\newcommand{\mbfhat}[1]{{\hat{\mbf{#1}}}}

\def\fdotb{{\raisebox{-0.6ex}{ \kern0.2ex\raisebox{0.8ex}{\tiny $\hspace*{-1ex}\circ$}}}}
\def\fddotb{{\raisebox{-0.6ex}{ \kern0.2ex\raisebox{0.8ex}{\tiny $\hspace*{-1ex}\circ\circ$}}}}
\newcommand{\ura}[1]{{\underrightarrow{{#1}}}}

\newcommand{\trans}{{\ensuremath{\mathsf{T}}}} 
\newcommand{\utimes}{ {\raisebox{-0.6ex}{ \kern-1.0ex\raisebox{0.6ex}{ \small $\mathsf{v}$}}} } %
\newcommand{\beq}{\begin{equation}}
\newcommand{\eeq}{\end{equation}}
\newcommand{\bdis}{\begin{displaymath}}
\newcommand{\edis}{\end{displaymath}}
\newcommand{\beqarray}{\begin{eqnarray}}
\newcommand{\eeqarray}{\end{eqnarray}}
\newcommand{\beqarraynn}{\begin{eqnarray*}}
\newcommand{\eeqarraynn}{\end{eqnarray*}}

\renewcommand{\theenumii}{\arabic{enumii}}

\makeatletter
\renewcommand{\p@enumii}{\theenumi.}
\makeatother

\let\OLDthebibliography\thebibliography
\renewcommand\thebibliography[1]{
  \OLDthebibliography{#1}
  \setlength{\parskip}{0.2mm}
  \setlength{\itemsep}{0.1pt plus 0.7ex}
}

\IEEEoverridecommandlockouts  
\ifCLASSINFOpdf
\else
\fi
\begin{document}
\input{cover.tex}
%
\title{Navigation and Control of Unconventional VTOL UAVs in Forward-Flight with Explicit Wind Velocity Estimation}
%
%
%

\author{Mitchell Cohen$^{1}$ and James Richard Forbes$^{2}$%
\thanks{Manuscript received: September 10th, 2019; Revised November 28th, 2019; Accepted January 7th, 2020.}
\thanks{This paper was recommended for publication by Editor Jonathan Roberts upon evaluation of the Associate Editor and Reviewers' comments.
This work was supported by MITACS Accelerate and NSERC Discovery Grants Program.} 
\thanks{$^{1}$Mitchell Cohen ({\tt\footnotesize mitchell.cohen3@mail.mcgill.ca}) and $^{2}$James Richard Forbes ({\tt\footnotesize james.richard.forbes@mcgill.ca}) are with the Department of Engineering, McGill University, Montreal QC, Canada, H3A 0C3}%
}
%
%

\markboth{IEEE Robotics and Automation Letters. Preprint Version. Accepted January, 2020}
{Cohen and Forbes: Navigation and Control of UAVs with Explicit Wind Velocity Estimation} 

%



\maketitle

\begin{abstract}
This paper presents a solution for the state estimation and control problems for a class of unconventional vertical takeoff and landing (VTOL) UAVs operating in forward-flight conditions. A tightly-coupled state estimation approach is used to estimate the aircraft navigation states, sensor biases, and the wind velocity. State estimation is done within a matrix Lie group framework using the Invariant Extended Kalman Filter (IEKF), which offers several advantages compared to standard multiplicative EKFs traditionally used in aerospace and robotics problems. An $SO(3)$-based attitude controller is employed, leading to a single attitude control law without a separate sideslip control loop. A control allocator is used to determine how to use multiple, possibly redundant, actuators to produce the desired control moments. The wind velocity estimates are used in the attitude controller and the control allocator to improve performance. A numerical example is considered using a sample VTOL tailsitter-type UAV with four control surfaces. Monte-Carlo simulations demonstrate robustness of the proposed control and estimation scheme to various initial conditions, noise levels, and flight trajectories.
\end{abstract}

\begin{IEEEkeywords}
Autonomous Vehicle Navigation, Sensor Fusion
\end{IEEEkeywords}

%
\IEEEpeerreviewmaketitle

\section{Introduction}
%
%
%
%
\IEEEPARstart{U}{nmaned} aerial vehicles (UAVs) are increasingly used for data collection, surveillance, delivery, and search and rescue missions. Vertical takeoff and landing (VTOL) UAVs leverage both vertical takeoff and efficient long-distance flight capabilities to realize a more versatile flight platform.  Recently, the use of VTOL UAVs has been explored by Uber, Amazon, Google, FedEx, and others \cite{bacchini2019electric}. Popular VTOL UAV configurations include tilt-rotors, tilt-wings, and tailsitters \cite{saeed2015review}. This paper considers control and state estimation of a tailsitter-type VTOL UAV operating in forward-flight conditions. A typical tailsitter initially ascends vertically in a similar way a multicopter would, before transitioning to a forward-flight configuration in order to fly like a traditional fixed-wing aircraft. VTOL UAVs may have additional lifting and control surfaces in unconventional configurations that are used during forward flight. These additional surfaces can be used to reconfigure the vehicle during flight \cite{alwi2008fault}. Alternatively, additional control surfaces realize a flight platform that is overactuated and thus robust to one or even multiple actuator failures \cite{alwi2008fault}. It then becomes necessary to determine how to correctly allocate the control moments to actuators in order to control the UAV in forward-flight. This is known as the control allocation problem, and several solutions are discussed in \cite{johansen2013control}. In some cases, the additional  unconventional control surfaces exert moments about multiple axes. Moreover, the moments generated are often highly dependant on the aircraft states and the wind velocity. As such, obtaining an estimate of the aircraft states, as well as the wind velocity, is important if reliable performance is to be realized. 

Several approaches exist for estimating the wind velocity during flight. For example, model-based approaches are explored in \cite{wenz2016combining} and \cite{lee2013estimation}, while a loosely-coupled approach is presented in \cite{cho2011wind}. In this paper, a tightly-coupled approach is employed where the wind velocity and navigation states are estimated using a single estimator.  In \cite{brossard2018tightly}, it was concluded that airspeed measuremetns improve the overall navigation solution through the correlation of the wind estimate and the velocity and attitude estimates. This paper builds on the work of \cite{brossard2018tightly}, where an EKF is considered, by deriving an ``Imperfect" Invariant Extended Kalman Filter (``Imperfect" IEKF) \cite{barrau2015non}.  The IEKF is a variant of the EKF that uses a very specific error definition motivated by the measurement model and matrix Lie group structure of the state-estimation problem \cite{barrau2016invariant,barrau2017three}. Moreover, as discussed in \cite{barrau2016invariant} and \cite{barrau2017three}, when certain conditions are met, the IEKF can be interpreted as an observer with attractive asymptotic stability properties. The ``Imperfect" IEKF derived in this paper provides an estimate of the navigation states, sensor biases, and the wind velocity. Due to the inclusion of sensor biases, and the specific form of the sensors, the state estimation problem considered herein does not satisfy the criterion of the IEKF. Namely, the group affine properties of the process model are not preserved in the presence of sensor biases, and a mix of left-, right-, and neither left-nor right-invariant sensors are used. Thus, the resultant state estimator is an ``Imperfect" IEKF, but still retains some of the attractive properties of the IEKF \cite{arsenault2019}. The state estimates generated are used in both the attitude control and control allocation loops. 

This paper explores the navigation and control of VTOL UAVs operating in forward-flight with multiple control surfaces and unconventional geometry. The contributions of this paper are, first, the adaptation of the geometric $SO(3)$-based attitude controllers commonly used for multicopters (in \cite{lee2010geometric} and \cite{mellinger2011minimum}, for example) for use with forward-flight aircraft in such a way that no explicit sideslip controller is needed to ensure balanced flight. The proposed controller uses the coordinated turn equation for forward-flight aircraft to ensure balanced flight, similar to \cite{oland2013quaternion}, but additionally uses the estimated airspeed within the controller. The second contribution of this paper is the use of an invariant filtering framework for the estimation of both the navigation states and wind velocity. The  combination  of  these two contributions  is  shown to  lead  to  reliable  performance  for  both  navigation  and control  of  unconventional  VTOL  UAVs operating in forward-flight conditions. Moreover, both the navigation and control strategies proposed herein can be used in the design of UAVs to ensure that new configurations can realize robust and reliable flight in the presence of wind. 

The remainder of the paper is structured as follows. Section~\ref{sec:eoms} presents the UAV equations of motion and Section~\ref{sec:aero} describes a model for their aerodynamics. Section~\ref{sec:scheme} describes the overall control and estimation strategy. Section~\ref{sec:IEKF} presents a state estimation filter in the invariant framework. Sections \ref{sec:alloc}, \ref{sec:attitude_control} present a solution for the control allocation and attitude control problem, respectively. Finally, Section~\ref{sec:sim} presents a numerical example with simulation results, while Section~\ref{sec:conclusion} provides concluding remarks.

%

\section{Equations of Motion} \label{sec:eoms}
A standard North-East-Down convention is used to define the basis vectors of $\mathcal{F}_a$, an inertial frame \cite{hughes2012spacecraft}. An unforced particle in $\mathcal{F}_a$ is denoted $w$ \cite{bernstein2008newton}. The frame that rotates with the aircraft body is denoted $\mathcal{F}_b$, and $\mbf{C}_{ba} \in SO(3)$ is the direction cosine matrix (DCM) that relates the attitude of $\mathcal{F}_b$ to the attitude of $\mathcal{F}_a$. The transpose of $\mbf{C}_{ba}$ is denoted $\mbf{C}_{ab}$, where $\mbf{C}_{ba} = \mbf{C}_{ab}^\trans$. A physical vector $\ura{v}$ can be resolved in either $\mathcal{F}_a$ or $\mathcal{F}_b$ as $\mbf{v}_a$ or $\mbf{v}_b$, and the relation between the two is $\mbf{v}_b = \mbf{C}_{ba} \mbf{v}_a$ or $\mbf{v}_a = \mbf{C}_{ab} \mbf{v}_b$.

A VTOL UAV in forward-flight is modelled as a rigid body with aerodynamic, gravitational and propulsion forces acting on it. Denoting point $z$ as the centre of mass of the body, the equations of motion are
	\begin{align} \nonumber
		m_\mathcal{B} \mbfdot{v}_a^{zw/a} = \mbf{C}_{ba}^\trans \mbf{f}_b^{\mathcal{B} z}, \hspace{5mm} \mbf{J}_b^{\mathcal{B} z} \mbsdot{\omega}_b^{ba} + \mbs{\omega}_b^{ba^\times} \mbf{J}_b^{\mathcal{B} z} \mbs{\omega}_b^{ba} = \mbf{m}_b^{\mathcal{B} z},
	\end{align}
where $m_\mathcal{B}$ is the mass of the aircraft, $\mbf{J}_b^{\mathcal{B} z}$ is the second moment of mass of the aircraft resolved in $\mathcal{F}_b$, $\mbs{\omega}_{b}^{ba}$ is the angular velocity of $\mathcal{F}_b$ relative to $\mathcal{F}_a$, resolved in $\mathcal{F}_b$. The cross operator $(\cdot)^\times$ is a mapping from $\mathbb{R}^3$ to $\mathfrak{so}(3)$ such that $\mbf{a}^\times \mbf{b} = - \mbf{b}^\times \mbf{a}, \hspace{2mm} \forall \mbf{a}, \mbf{b} \in \mathbb{R}^3$. In addition, $\mbf{f}_b^{\mathcal{B} z}$ and $\mbf{m}_b^{\mathcal{B} z}$ represent the forces and moments acting on the body, resolved in $\mathcal{F}_b$.
The forces acting on the aircraft are $\mbf{f}_b^{\mathcal{B} z} = \mbf{f}_b^p + \mbf{f}_b^a + \mbf{C}_{ba} \mbf{f}_a^g$,
where $\mbf{f}_a^g = \begin{bmatrix} 0 & 0 & m_\mathcal{B} g \end{bmatrix}^\trans$ is the gravitational force resolved $\mathcal{F}_a$ and $g = 9.81 \mathrm{m}/\mathrm{s}^2$, $\mbf{f}_b^p = \begin{bmatrix} T & 0 & 0 \end{bmatrix}^\trans$ represents propulsion forces, where $T$ is a thrust force, and $\mbf{f}_b^a$ represents aerodynamic forces acting on the UAV.
The moments acting on the aircraft are aerodynamic moments due to both fixed and movable aerodynamics surfaces, denoted $\mbf{m}_b^a$, and thus $\mbf{m}_b^{\mathcal{B} z} = \mbf{m}_b^a$.

The equations of motion are completed by the translational and rotational kinematics, respectively given by $\mbfdot{r}_a^{zw} = \mbf{v}_a^{zw/a}$ and $\mbfdot{C}_{ab} = \mbf{C}_{ab} \mbs{\omega}_{b}^{ba^\times}$, where $\mbf{r}_a^{zw}$ is the position of the UAV.

\section{Aerodynamic Modelling} \label{sec:aero}
To model the forces and moments generated by the aerodynamic surfaces, a component breakdown approach is used \cite{Khan2016, Khan2016_thesis}. In this approach, the aircraft is split up into segments, and each segment produces a lift and a drag force at its aerodynamic centre. The forces and the moments about point $z$ due to the applied forces are then summed.

Denote particle $q$ to be a particle moving with the wind field around the UAV. The velocity vector of the wind is given by $\ura{v}^{qw/a}$ \cite{ansari2017retrospective}. Denote the aerodynamic centre of the $\imath$'th segment as $c_\imath$ and the position of $c_\imath$ relative to $z$, resolved in $\mathcal{F}_b$, as $\mbf{r}_b^{c_\imath z}$. Define the segment frame as $\mathcal{F}_{d_\imath}$, that rotates with the aerodynamic segment. The DCM that relates the attitude of the segment frame to the attitude of the body frame is given by $\mbf{C}_{d_\imath b}$. If the particular surface that is being modelled is a control surface, the DCM $\mbf{C}_{d_\imath b}$ is a function of the control surface deflection denoted $\delta_\imath$.

The velocity of the segment relative to the surrounding air resolved in $\mathcal{F}_b$ is $\mbf{v}_{b}^{c_\imath q/a} = \mbf{C}_{ba} \mbf{v}_a^{zw/a} + \mbs{\omega}_b^{ba^\times} \mbf{r}_b^{c_\imath z} - \mbf{C}_{ba} \mbf{v}_a^{qw/a}$.
This velocity can then be resolved in the segment frame as $\mbf{v}_{d_\imath}^{c_\imath q/a} = \mbf{C}_{d_\imath b} \mbf{v}_b^{c_\imath q/a}$ and can be used to calculate the segment angle of attack, denoted $\alpha_\imath$, and the segment sideslip angle, denoted $\beta_\imath$, using 
	\begin{align} \nonumber
		\alpha_{\imath} = \mathrm{atan}2 \left(v_{d_\imath,3}^{c_\imath q/a}, v_{d_\imath ,1}^{c_\imath q/a} \right), \hspace{4mm} \beta_{\imath} = \sin^{-1} \left(\frac{v_{d_\imath ,2}^{c_\imath q/a}}{\norm{\mbf{v}_{d_\imath}^{c_\imath q/a}}} \right).
	\end{align}
The segment angle of attack and segment sideslip angle are then used to define the segment stability frame and wind frame. The DCM that relates the segment's stability frame to the segment frame is given by $\mbf{C}_{s_\imath d_\imath} = \mbf{C}_2(-\alpha_\imath)$,
where $\mbf{C}_2(\cdot)$ is the DCM describing the second principal rotation. The DCM that relates the segment wind frame, $\mathcal{F}_{w_\imath}$, to the stability frame is given by $\mbf{C}_{w_\imath s_\imath} = \mbf{C}_3(\beta_{s_\imath})$. \\
By definition, lift and drag forces of each segment, denoted $\ura{f}^{L_\imath}$ and $\ura{f}^{D_\imath}$ respectively, act along the axes of the wind frame of that segment such that $\ura{f}^{D_\imath} = -f^{D_\imath} \ura{w_\imath}^1$ and  $\ura{f}^{L_\imath} = -f^{L_\imath} \ura{w_\imath}^3$ \cite{yan2011longitudinal}.
The aerodynamic forces of the $\imath$'th segment resolved in the wind frame are given by 
	\begin{align} \nonumber
		\mbf{f}_{w_\imath}^{\, a_\imath} = \begin{bmatrix} -f_{w_\imath}^{D_\imath} & 0 & -f_{w_\imath}^{L_\imath} \end{bmatrix}^\trans,
	\end{align}
where lift and drag forces are given by a flat plate model as 
	\begin{align}  \nonumber \scriptsize
		f_w^{L_\imath} & = \frac{1}{2} \rho_\mathrm{air} {\norm{\mbf{v}_a^{c_\imath q/a}}}^2 S_\imath C_L(\alpha_\imath), \\ \nonumber
		f_w^{D_\imath} & = \frac{1}{2} \rho_\mathrm{air} {\norm{\mbf{v}_a^{c_\imath q/a}}}^2 S_\imath C_D(\alpha_\imath),
	\end{align}
where $\rho_\mathrm{air}$ is the density of air and is assumed to be constant, $S_\imath$ is the surface area of the $\imath$'th surface, and lift and drag coefficients $C_L$ and $C_D$ are modelled as a function of the angle of attack.
The aerodynamic forces are then resolved back in $\mathcal{F}_b$ though
	\begin{align}  \nonumber
		\mbf{f}_b^{\, a_\imath} = \mbf{C}_{d_\imath b}^\trans  \mbf{C}_{s_\imath d_\imath}^\trans \mbf{C}_{w_\imath s_\imath}^\trans \mbf{f}_w^{\, a_\imath}.
	\end{align} 
The aerodynamic moments generated by each segment are then summed up using
	\begin{align} \nonumber
		\mbf{m}_b^{a} = \sum_{\imath=1}^{n}  \mbf{r}_b^{c_\imath z^\times} \mbf{f}_{b}^{\, a_\imath}.
	\end{align}
It follows that the moments, $\mbf{m}_b^a$, are a nonlinear function of the aircraft states, the wind velocity, and control surface deflections $\mbs{\delta} = \begin{bmatrix} \delta_1 & \delta_2 & \ldots & \delta_n \end{bmatrix}^\trans$, where $n$ is the total number of control surfaces, such that 
	\begin{align} \label{eq:mapping}
		\mbf{m}_b^a = \mbf{f} \left(\mbf{C}_{ba}, \mbs{\omega}_b^{ba}, \mbf{v}_a^{zw/a}, \mbf{v}_a^{qw/a}, \mbs{\delta} \right).
	\end{align}
\section{Control Objectives and Overall Architecture} \label{sec:scheme}
\begin{figure}[H]
			\centering
        	\includegraphics[width=0.45\textwidth]{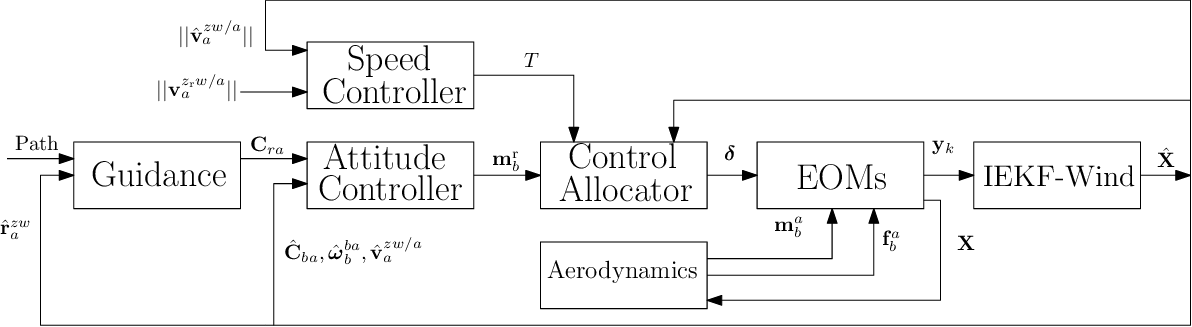}
		\caption{Overall control architecture.}
		\label{fig:control_scheme}
		\end{figure}

The objective at hand is to control a VTOL UAV operating in forward-flight with an arbitrary amount of control surfaces along a path at a given velocity. The problem is split into the following stages, all shown in Figure~\ref{fig:control_scheme}.
	\begin{enumerate}
		\item \textit{State Estimation: } The UAV navigation states, along with sensor biases and wind velocity, are estimated using an ``Imperfect" IEKF. The estimated states are then used within the controller and control allocator.
		\vspace{-2mm}
		\item \textit{Guidance: } The guidance stage determines how to orient the UAV so that its position converges to the path. Denoting the desired reference frame of the UAV by $\mathcal{F}_r$, the guidance law outputs the DCM $\mbf{C}_{ra}$.
		\vspace{-2mm}
		\item \textit{Attitude Control: } An attitude controller ensures convergence of the aircraft attitude to the desired aircraft attitude $\mbf{C}_{ra}$, while also ensuring that balanced flight with zero sideslip is achieved. The attitude controller outputs control moments $\mbf{m}_b^{\mathrm{r}}$ to track $\mbf{C}_{ra}$.
		\vspace{-2mm}
		\item \textit{Speed Control: } A speed controller generates a required thrust force to ensure that the magnitude of the aircraft velocity converges to a desired aircraft velocity. 
		\vspace{-2mm}
		\item \textit{Control Allocation: } A control allocator determines how to generate the required control moment output by the attitude controller using onboard actuators.
		\vspace{-2mm}
	\end{enumerate}
		\vspace{-5mm}
Note that the proposed navigation, guidance, and attitude control solutions are directly applicable to a wide class of VTOL UAVs in forward-flight conditions. The specific form of the control allocator is platform dependant and depends on the configuration of the onboard actuators.

\section{``Imperfect" IEKF-Wind Estimator} \label{sec:IEKF}
A general matrix Lie group is composed of invertible $n \times n$ matrices with $k$ degrees of freedom closed under matrix multiplication \cite{hall2013lie}. The aircraft navigation states, namely the aircraft attitude, velocity, and position can be cast into an element of the matrix Lie group of double direct isometries, $SE_2(3)$ \cite{eade2014lie}. However, sensor biases and wind velocity are not elements of any traditional matrix Lie group. It is still possible to use an invariant filter framework in this case by deriving an ``Imperfect" IEKF, as done in \cite{barrau2015non} and \cite{heo2018consistent}. In the ``Imperfect" IEKF, the formulation of the problem leads to multiplicative error terms for states defined on a matrix Lie group and additive error terms for states defined on a linear vector space.
\subsection{Process Model}
Consider biased, noisy rate-gyro measurements given by $\mbf{u}_b^1 = \mbs{\omega}_b^{ba} - \mbs{\beta}_b^1 - \mbf{w}_b^1$, where $\mbf{w}_b^1 \sim \mathcal{N}(\mbf{0}, \mbf{Q}_1)$ and $\mbs{\beta}_b^1$ is the bias. Bias is modelled as a random walk, $\dot{\mbs{\beta}}_b^1 = \mbf{w}_b^3$, where $\mbf{w}_b^3 \sim \mathcal{N}(\mbf{0}, \mbf{Q}_4)$. \\ 
Consider biased, noisy accelerometer accelerometer measurements given by $\mbf{u}_b^2 = \mbf{f}_b - \mbs{\beta}_b^2 - \mbf{w}_b^2$, where $\mbf{w}_b^2 \sim \mathcal{N}(\mbf{0}, \mbf{Q}_2)$ and $\mbf{f}_b$ is the specific force vector resolved in the body frame. In addition, the accelerometer bias $\mbs{\beta}_b^2$ is also modelled as a random walk and thus $\dot{\mbs{\beta}}_b^2 = \mbf{w}_b^4$, where $\mbf{w}_b^4 \sim \mathcal{N}(\mbf{0}, \mbf{Q}_5)$. \\
The acceleration can then be expressed as 
	\begin{align} \nonumber
		\mbfdot{v}_a^{zw/a} = \mbf{C}_{ab} (\mbf{u}_b^2 + \mbs{\beta}_b^2 + \mbf{w}_b^2) + \frac{\mbf{f}_a^g}{m_\mathcal{B}}.
	\end{align}	
The velocity of a wind particle $q$, resolved in the inertial frame, is denoted $\mbf{v}_a^{qw/a}$. The wind is modelled as a random walk process with $\mbfdot{v}_a^{qw/a} = \mbf{w}_a^1$,
where $\mbf{w}_a^1 \sim \mathcal{N}(\mbf{0}, \mbf{Q}_3)$.
Thus, the continuous-time kinematic process model $\mbfdot{X} = \mbf{F}(\mbf{X}, \mbf{u}, \mbf{w})$, where $\mbf{X}$ is an element of a matrix Lie group, is given by
	\begin{align} \label{eq:process_model_1}
			\dot{\mbf{C}}_{ab} & = \mbf{C}_{ab} (\mbf{u}_b^1 + \mbs{\beta}_b^1 + \mbf{w}_b^1)^\times,  \\ 
		\dot{\mbf{r}}_a^{zw} & = \mbf{v}_a^{zw/a}, \\ 
		\dot{\mbf{v}}_a^{zw/a} & = \mbf{C}_{ab} (\mbf{u}_b^2 + \mbs{\beta}_b^2 + \mbf{w}_b^2) + \frac{\mbf{f}_a^g}{m_\mathcal{B}}, \\ 
		\dot{\mbs{\beta}}_b^1 & = \mbf{w}_b^3, \hspace{5mm}\dot{\mbs{\beta}}_b^2 = \mbf{w}_b^4, \hspace{5mm} \mbfdot{v}_a^{qw/a} = \mbf{w}_a^1. \label{eq:process_model_2}
	\end{align}
Note that \eqref{eq:process_model_1} to  \eqref{eq:process_model_2} together constitute the continuous-time process model.
The aircraft navigation states can be cast into an element of the matrix Lie group of double direct isometries, $SE_2(3)$. Denoting the matrix $\mbf{Y}$ as an element of $SE_2(3)$, the states $\mbf{C}_{ab}, \mbf{v}_a^{zw/a}$, and $\mbf{r}_a^{zw}$ can be placed into an element of $SE_2(3)$ as
	\begin{align} \nonumber
		\mbf{Y} = 
		\begin{bmatrix} 
			\mbf{C}_{ab} & \mbf{v}_a^{zw/a} & \mbf{r}_a^{zw} \\ 
			 & 1 &  \\ 
			 & & 1 
		\end{bmatrix} \in SE_2(3),
	\end{align}
where non-essential zero entries are omitted. \\ 
Similar to \cite{heo2018consistent}, the entire state including biases and wind velocity can be placed into an element of a new matrix Lie group $\mathcal{G}$. An element of $\mathcal{G}$ is written as 
	\begin{align} \nonumber
		\mbf{X} = 
		\begin{bmatrix} 
			\mbf{Y} & \mbf{0} & & & \\ 
			& \mbf{1} & \mbf{v}_a^{qw/a} & \mbs{\beta}_b^1 & \mbs{\beta}_b^2 \\
			 & & 1 & & \\
			 & & & 1 & \\
			 & & & & 1 
		\end{bmatrix} \in \mathcal{G}.
	\end{align}
The inverse of an element of $\mathcal{G}$ is then defined such that $\mbf{X} \mbf{X}^{-1} = \mbf{1}$, where $\mbf{1}$ is the identity element.
Let $\mathfrak{g}$ be the matrix Lie algebra of $\mathcal{G}$. The matrix Lie algebra $\mathfrak{g}$ is the tangent space of $\mathcal{G}$ at the identity element. The operator $(\cdot)^\vee : \mathfrak{g} \to \mathbb{R}^k$ maps the matrix Lie algebra to a $k$-dimention column matrix and the inverse map is defined $(\cdot)^\wedge : \mathbb{R}^k \to \mathfrak{g}$. The column matrix $\mbs{\xi} \in \mathbb{R}^{18}$ is mapped to $\mathfrak{g}$ using
	\begin{align} \nonumber
		\mbs{\xi}^\wedge = \begin{bmatrix} \mbs{\xi}^{Y} \\ \mbs{\xi}^\mathrm{w} \\ \mbs{\xi}^{\beta^1} \\ \mbs{\xi}^{\beta^2} \end{bmatrix}^\wedge = \begin{bmatrix} \mbs{\xi}^{Y^\wedge} & \mbf{0} & & & \\ & \mbf{0} & \mbs{\xi}^\mathrm{w} & \mbs{\xi}^{\beta^1} & \mbs{\xi}^{\beta^2} \\ & & & & \mbf{0} \end{bmatrix} \in \mathfrak{g},
	\end{align}
where $\mbs{\xi}^{Y} = \begin{bmatrix} \mbs{\xi}^{\phi^\trans} & \mbs{\xi}^{\mathrm{v}^\trans} & \mbs{\xi}^{\mathrm{r}^\trans} \end{bmatrix}^\trans.$
The mapping between the matrix Lie algebra and the matrix Lie group is the exponential map, defined by the matrix exponential, $\mathrm{exp} (\cdot):~ \mathfrak{g} \to \mathcal{G}$. The exponential map from $\mathfrak{g}$ to $\mathcal{G}$ is given by
	\begin{align} \nonumber
		\mathrm{exp}(\mbs{\xi}^\wedge) = 
			\begin{bmatrix} \mathrm{exp}_{SE_2(3)} \left(\mbs{\xi}^{Y^\wedge} \right) & \mbf{0} & & & \\
			 & \mbf{1} & \mbs{\xi}^\mathrm{w} & \mbs{\xi}^{\beta^1} & \mbs{\xi}^{\beta^2} \\
			 & & 1 & & \\
			 & & & 1 & \\
			 & & & & 1 \end{bmatrix} \in \mathcal{G}.
	\end{align}
\subsection{Measurement Model}
The available measurements are assumed to be discrete GPS position and velocity measurements, written $\mbf{y}_{a_k}^1$ and $\mbf{y}_{a_k}^2$ respectively, as well as discrete pitot tube measurements and magnetometer measurements.  The GPS measurements can be written as a function of the navigation states contained within $\mbf{Y}$ to yield a left-invariant measurement model of the form
	\begin{align} \nonumber
		\begin{bmatrix}
			\mbf{y}_{a_k}^1 \\
			0 \\
			1 \\
			\mbf{y}_{a_k}^2 \\
			0 \\
			1
		\end{bmatrix}
		=
		\begin{bmatrix}
			\mbf{Y}_{k}
			\begin{bmatrix}
				\mbf{0} \\
				1
			\end{bmatrix}
			+
			\begin{bmatrix}
				\mbf{v}_{a_k}^1 \\
				\mbf{0}
			\end{bmatrix} \\[2ex]
			\mbf{Y}_{k}
			\begin{bmatrix}
				\mbf{0} \\
				1 \\
				0
			\end{bmatrix}
			+
			\begin{bmatrix}
				\mbf{v}_{a_k}^2 \\
				\mbf{0}
			\end{bmatrix}
		\end{bmatrix},
	\end{align}
where $\mbf{v}_{a_k}^1 \sim \mathcal{N}(\mbf{0}, \mbf{R}_1)$ and $\mbf{v}_{a_k}^2 \sim \mathcal{N}(\mbf{0}, \mbf{R}_2)$. The GPS measurement models are left-invariant as they are in the form $\mbf{y}_k^\mathrm{L} = \mbf{X}_k \mbf{b}_k + \mbf{v}_k$
where $\mbf{X}_k$ is an element of a matrix Lie group, $\mbf{b}_k$ is some known vector and $\mbf{v}_k$ is white, Gaussian noise.

The pitot tube measures the component of $\ura{v}^{zq/a}$ along the $\ura{b}^1$ axis. Thus, the measurement model for  pitot tube measurements is given by
\vspace{-1mm} 
		\begin{align} \label{eq:pitot_measure}
		y_{b_k} = \mbf{1}_1^{\trans} \mbf{C}_{a b_{k}}^\trans \left(\mbf{v}_a^{z_k w/a} - \mbf{v}_a^{q_k w/a} \right) + v_{b_k}^1,
	\end{align}
where $\mbf{1}_1^{\trans}$ is the first column of the identity matrix and $v_{b_k}^1 \sim \mathcal{N}(0, R_3)$. Note that \eqref{eq:pitot_measure} is neither left- nor right-invariant. \\
The right-invariant measurement model for the magnetometer is written
	\begin{align} \nonumber
		\mbf{y}_{b_k}^2 = \mbf{C}_{a b_k}^\trans \mbf{m}_a + \mbf{v}_{b_k}^2,
	\end{align}
where $\mbf{m}_a$ is the magnetic field vector resolved in $\mathcal{F}_a$ and $\mbf{v}_{b_k}^2 \sim \mathcal{N}\left(\mbf{0}, \mbf{R}_4 \right)$. The magnetometer measurement model is said to be right-invariant since it is of the form $\mbf{y}_k^\mathrm{R} = \mbf{X}_k^{-1} \mbf{b}_k + \mbf{v}_k$, where, similarly to the left-invariant measurement model, $\mbf{X}_k$ is an element of a matrix Lie group, $\mbf{b}_k$ is a known vector, and $\mbf{v}_k$ is Gaussian, white noise.
\subsection{IEKF Equations}
To derive the invariant filter, a left-invariant error definition will be used, written $\delta \mbf{X} = \mbf{X}^{-1} \mbfhat{X}$. This left-invariant error will be used when computing the Jacobians associated with the linearized process model and the innovation. This error is said to be left-invariant because the error is invariant to left multiplication by an element of $\mathcal{G}$ \cite{bonnable2009invariant}. The left-invariant error is used because there are two left-invariant measurements, the GPS position and velocity measurements, while there is one right-invariant measurement, the magnetometer, and one neither left- nor right-invariant measurement, the pitot tube. For the group $\mathcal{G}$, the left-invariant error $\delta \mbf{X}$ can be expanded as
	\begin{align} \nonumber
		\delta \mbf{X} = \begin{bmatrix} \mbf{Y}^{-1} \mbfhat{Y} & \mbf{0} & & & \\ & \mbf{1} & \mbfhat{v}_a^{qw/a} - \mbf{v}_a^{qw/a} & \hat{\mbs{\beta}}_b^1 - \mbs{\beta}_b^1 & \hat{\mbs{\beta}}_b^2 - \mbs{\beta}_b^2 \\ & & 1 & & \\ & & & 1 & \\ & & & & 1 \end{bmatrix}.
	\end{align}
The subblocks of $\delta \mbf{X}$ allow for the definition of the state errors
	\begin{align} \nonumber
		\delta \mbf{Y} & = \mbf{Y}^{-1} \mbfhat{Y}, \hspace{7mm} \delta \mbf{v}_a^{qw/a} = \mbfhat{v}_a^{qw/a} - \mbf{v}_a^{qw/a}, \\ \nonumber
		\delta \mbs{\beta}_b^1 & = \hat{\mbs{\beta}}_b^1 - \mbs{\beta}_b^1,  \hspace{5mm}\delta \mbs{\beta}_b^2 = \hat{\mbs{\beta}}_b^2 - \mbs{\beta}_b^2.
	\end{align}
Expanding the left-invariant error for the aircraft navigation states on $SE_2(3)$, $\delta \mbf{Y} = \mbf{Y}^{-1} \mbfhat{Y}$, yields
	\begin{align} \nonumber
		\delta \mbf{C} & = \mbf{C}_{ab}^\trans \hat{\mbf{C}}_{ab}, \hspace{2mm}
		\delta \mbf{v} = \mbf{C}_{ab}^\trans \left(\hat{\mbf{v}}_a^{zw/a} - \mbf{v}_a^{zw/a} \right), \\ \nonumber
		\delta \mbf{r} & = \mbf{C}_{ab}^\trans \left(\hat{\mbf{r}}_a^{zw} - \mbf{r}_a^{zw} \right).
	\end{align}
These error definitions, which are different than those used in \cite{brossard2018tightly}, are used to linearize the process model and innovation. Linearizing the process model gives
	\begin{align} \nonumber
		\delta \mbsdot{\xi} = \mbf{A} \delta \mbs{\xi} + \mbf{L} \delta \mbf{w},
	\end{align}
where the $\mbf{A}$ and $\mbf{L}$ matrices are given by 
	\begin{align} \nonumber
	\scriptsize
		\mbf{A} = 
		\begin{bmatrix} 
			(-\mbf{u}_b^1 - \mbshat{\beta}_b^1)^\times & \mbf{0} &  &  & \mbf{1} &  \\
		 -\left(\mbf{u}_b^2 + \mbshat{\beta}_b^2 \right)^\times & -\left(\mbf{u}_b^1 + \mbshat{\beta}_b^1 \right)^\times &  & &  & \mbf{1} \\ 
		  & \mbf{1} & -\left(\mbf{u}_b^1 + \mbshat{\beta}_b^1 \right)^\times &  &  & \\ 
		  & & & \mbf{0} & & \\
		  & & & & \mbf{0} &  \\ 
		  & & & & & \mbf{0} \end{bmatrix},
	\end{align}
	\begin{align} \nonumber
		\mbf{L} = -\mbf{1}.
	\end{align}
Note that due to the way the errors have been defined in the invariant framework, the process model Jacobians only depend on $\mbf{u}_b^1$ and $\mbf{u}_b^2$ and the bias estimates. These Jacobians are less dependant on the state estimate than the ones derived with the MEKF in \cite{brossard2018tightly}, and this is the advantage of the ``Imperfect" IEKF over the MEKF, as poor state estimates can lead to inaccurate Jacobians. The Jacobians derived using the invariant framework make the ``Imperfect" IEKF less suceptible to initialization errors and leads to better performance in the transient compared to the MEKF \cite{arsenault2019}.
The linearized process model is then discretized using any appropriate method. The state and covariance prediction steps are then
	\begin{align} \nonumber
		\check{\mbf{X}}_k & = \mbf{F}_{k-1} (\mbfhat{X}_{k-1}, \mbf{u}_{k-1}), \\  \nonumber
		\check{\mbf{P}}_k & = \mbf{A}_{k-1} \hat{\mbf{P}}_{k-1} \mbf{A}_{k-1}^\trans + \mbf{L}_{k-1} \mbf{Q}_{k-1} \mbf{L}_{k-1}^\trans,
	\end{align}
where $\mbf{F}_{k-1}$ is the discrete-time process model. In forthcoming simulations, a forward-Euler discretization scheme is used to discretize the continuous-time process model given by \eqref{eq:process_model_1} to \eqref{eq:process_model_2}.
The state correction is then given by
	\begin{align} \nonumber
		\mbfhat{X}_k = \check{\mbf{X}}_k \mathrm{exp} \left(- (\mbf{K}_k \mbf{z}_k)^\wedge \right),
	\end{align}
where $\mbf{z}_k$ is the innovation and the Kalman gain $\mbf{K}_k$ is computed using $\mbf{K}_k = \check{\mbf{P}}_k \mbf{H}_k^\trans \left(\mbf{H}_k \check{\mbf{P}}_{k} \mbf{H}_k^\trans + \mbf{M}_k \mbf{R}_k \mbf{M}_k^\trans \right)^{-1}$.
The covariance is updated using
	\begin{align}\nonumber
	\hat{\mbf{P}}_k & = \left(\mbf{1} - \mbf{K}_k \mbf{H}_k \right) \check{\mbf{P}}_k \left(\mbf{1} - \mbf{K}_k \mbf{H}_k \right)^\trans + \mbf{K}_k \mbf{M}_k \mbf{R}_k \mbf{M}_k^\trans \mbf{K}_k^\trans.
	\end{align}
The linearized innovation Jacobians, with respect to the states and measurement noise, are given by 
	\begin{align} \tiny \nonumber
			\mbf{H}_{k} = 
			\begin{bmatrix} \mbf{0} & \mbf{0} & -\mbf{1} & \mbf{0} & \mbf{0} & \mbf{0} \\ 
			\mbf{0} & -\mbf{1} & \mbf{0} & \mbf{0} & \mbf{0} & \mbf{0} \\  
			-\mbf{1}_1^\trans \big( (\check{\mbf{C}}_{a b_k}^\trans \check{\mbf{v}}_a^{z_k w/a})^\times - (\check{\mbf{C}}_{a b_k}^\trans \check{\mbf{v}}_a^{q_k w/a} )^\times \big)  & -\mbf{1}_1^\trans & \mbf{0} & \mbf{1}_1^\trans \check{\mbf{C}}_{ab}^\trans & \mbf{0} & \mbf{0} \\ 
			-\left(\check{\mbf{C}}_{a b_k}^\trans \mbf{m}_a \right)^\times & \mbf{0} & \mbf{0} & \mbf{0} & \mbf{0} & \mbf{0} \end{bmatrix}
	\end{align}
	\vspace{-2mm}
	\begin{align} 	\small \nonumber
		\mbf{M}_k = \mathrm{diag} \left(\check{\mbf{C}}_{ab}^\trans, \check{\mbf{C}}_{ab}^\trans, 1, \mbf{1} \right).
	\end{align}
	\vspace{-8mm}
	
\section{Control Allocator} \label{sec:alloc}
The purpose of the control allocator is to determine the required control surface deflections to generate a desired moment $\mbf{m}_{b}^\mathrm{r}$. The problem can be transformed into a linear problem by linearizing \eqref{eq:mapping} about small control surface deflections. Denote $\mbfhat{B}$ as the Jacobian of \eqref{eq:mapping} with respect to $\mbs{\delta}$ evaluated at the estimate of the aircraft states $\mbfhat{C}_{ba}$, $\mbshat{\omega}_b^{ba}$, $\mbfhat{v}_a^{zw/a}$ and $\mbfhat{v}_b^{qw/a}$. The control allocation problem then involves finding the control surface deflections such that $\mbf{m}_b^{\mathrm{r}} = \mbfhat{B} \mbs{\delta}$. Unconstrained allocation techniques involve generalized pseudoinverses, while constrained techniques such as the direct control allocation are detailed in \cite{Oppenheimer2006}.  In this paper, pseudoinverse methods are used to calculate the control surface deflections using $\mbs{\delta} = \mbfhat{B}^\dagger \mbf{m}_b^\mathrm{r}$, where $(\cdot)^\dagger$ denotes the pseudoinverse. Note that because the Jacobian $\mbfhat{B}$ is evaluated at the best estimate of the aircraft states, $\mbf{m}_b^a$ will not be exactly equal to $\mbf{m}_b^\mathrm{r}$, but only approximately so.

\section{Attitude Controller} 
\label{sec:attitude_control}
The goal of the attitude controller is to yield convergence of the aircraft attitude to some reference attitude, while ensuring that balanced flight is achieved, meaning that the second component of the airspeed resolved in $\mathcal{F}_b$ must be zero (i.e., $v_{b,2}^{zq/a} = 0$). The reference DCM, $\mbf{C}_{ra}$, can be constructed using a 3-2-1 Euler angle sequence as $\mbf{C}_{ra} = \mbf{C}_1(\phi_\mathrm{r}) \mbf{C}_2 (\theta_\mathrm{r}) \mbf{C}_3(\psi_\mathrm{r})$, where $\phi_\mathrm{r}$, $\theta_\mathrm{r}$, and $\psi_\mathrm{r}$ are reference roll, pitch and yaw angles. 
Given a desired roll angle trajectory $\phi_\mathrm{r}$, the corresponding desired yaw angle rate $\dot{\psi}_\mathrm{r}$ trajectory can be determined from the coordinated turn equation  for fixed-wing aircraft given by \cite{beard2012small} 
	\begin{align} \nonumber
		\dot{\psi}_\mathrm{r} = \frac{g}{\norm{\mbfhat{v}_a^{zq/a}}} \tan \phi_\mathrm{r}.
	\end{align}
Note that the estimated airspeed is also used in the attitude controller to ensure balanced flight.
The desired Euler angle rates, $\mbsdot{\Theta}_\mathrm{r} = \begin{bmatrix} \dot{\phi}_\mathrm{r} & \dot{\theta}_\mathrm{r} & \dot{\psi}_\mathrm{r} \end{bmatrix}^\trans$, can be used to compute a desired angular velocity using the mapping between Euler angle rates and angular velocity given by \cite{hughes2012spacecraft}
	\begin{align}	 \nonumber
		\mbs{\omega}_r^{ra} = \begin{bmatrix} \mbf{1}_1 & \mbf{C}_1(\phi_\mathrm{r}) \mbf{1}_2 & \mbf{C}_1(\phi_\mathrm{r}) \mbf{C}_2(\theta_\mathrm{r}) \mbf{1}_3 \end{bmatrix}\mbsdot{\Theta}_\mathrm{r}.
	\end{align}
The angular velocity error resolved in $\mathcal{F}_b$ is then written as $\mbf{e}_b^{\mbs{\omega}} = \mbs{\omega}_b^{ba} - \mbf{C}_{br} \mbs{\omega}_{r}^{ra}$.
Consider the attitude error DCM given by $\mbf{C}_{br} = \mbf{C}_{ba} \mbf{C}_{ra}^\trans$. A measure of attitude error is chosen as $\mbs{\phi}^e = \frac{1}{2} \left(\mbf{C}_{br} - \mbf{C}_{br}^\trans \right)^\vee$.
The attitude controller given by  \cite{lee2010geometric}
	\begin{align} \label{eq:attitude_controller}
		\mbf{m}_b^\mathrm{r} = \mbf{K}^\mbs{\phi} \mbs{\phi}^e - \mbf{K}^\mbs{\omega} \mbf{e}_b^{\mbs{\omega}} + \mbf{K}^{i,1} \int_0^t \left(\mbf{K}^{i,2} \mbs{\phi}^e - \mbf{e}_b^\mbs{\omega}\right) \mathrm{d} \tau,
	\end{align}
is then used to generate control torques to control the orientation of the UAV, where $\mbf{K}^\mbs{\phi}$, $\mbf{K}^\mbs{\omega}$, $\mbf{K}^{i,1}$, and $\mbf{K}^{i,2}$ are symmetric positive definite gain matrices. The article \cite{lee2010geometric} provides insight on the selection of gains for stability purposes. Note that the control law of \eqref{eq:attitude_controller} uses the DCM directly,  rather than using a parametrization of the attitude such as Euler angles or quaternions. Doing so avoids the deficiencies of attitude parametrizations, such as not being a unique or global representation of attitude \cite{mellinger2011minimum}.

In this method of attitude control, a separate sideslip controller, as done in \cite{beard2012small}, is not needed since the desired reference frame and angular velocity have been chosen to satisfy the coordinated turn equation.  The coordinated turn equation is used in \cite{oland2013quaternion} in a similar way. However, \cite{oland2013quaternion} assumes that the true airspeed is known and available, which is not the case in practice. Moreover, \cite{oland2013quaternion} uses a different guidance and control interconnection, as well as a quaternion representation of attitude, unlike the proposed controller.
\vspace{-2mm}

\section{Numerical Example and Simulation Results} \label{sec:sim}
\begin{figure}[H]
			\centering
        		\includegraphics[width=0.28\textwidth]{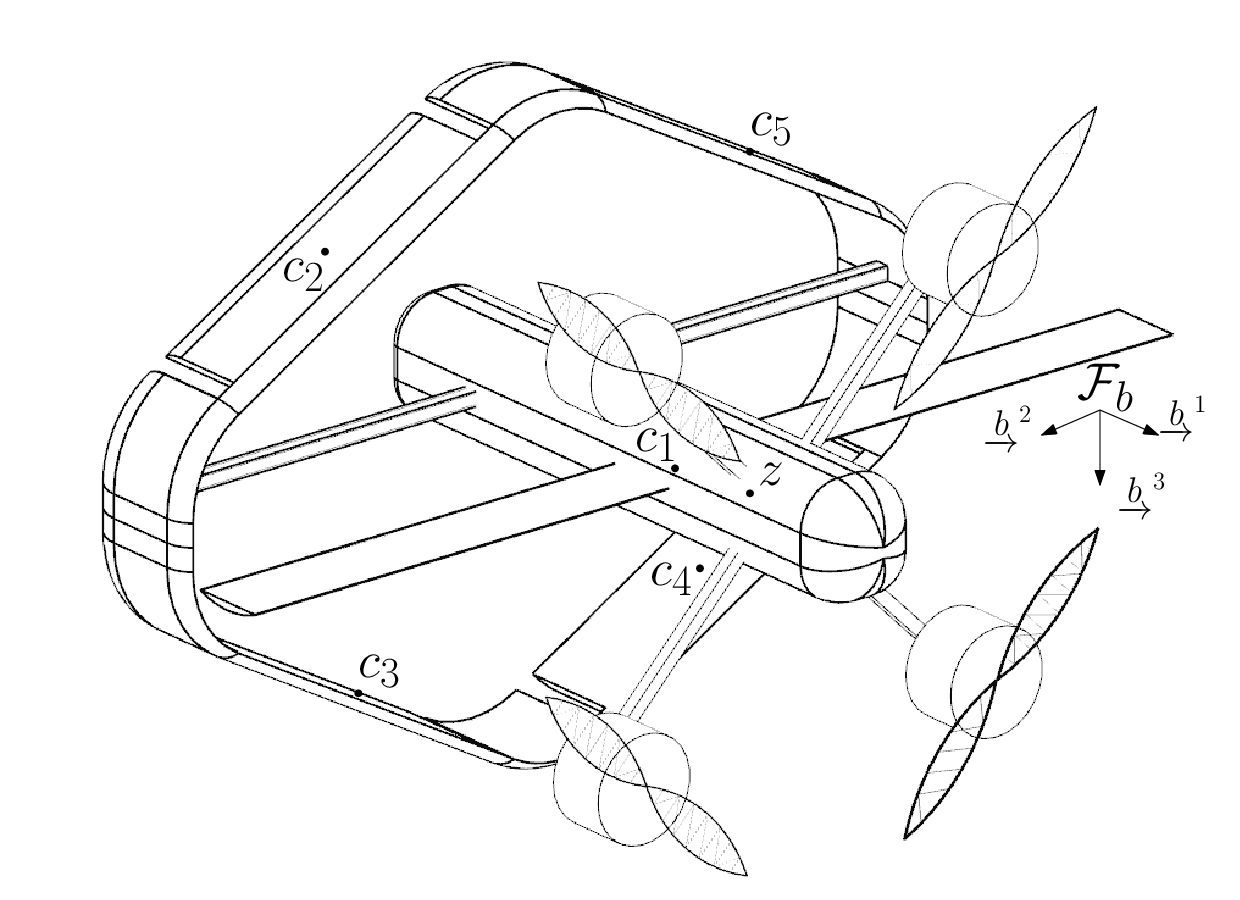}
        		\vspace{-6mm}
        \caption{Sample aircraft with unconventional control surfaces.}
		\label{fig:aircraft_setup}
		\end{figure}

The VTOL tailsitter-type UAV considered for the numerical example has one primary airfoil to provide lift, and four control surfaces arranged in a double inverted V-Tail type configuration, as shown in Figure~\ref{fig:aircraft_setup}.
The aerodynamic centre of the primary airfoil is denoted $c_1$, and the aerodynamic centres of the control surfaces are denoted $c_\imath$, $\imath = 2,3,4,5$. The control surfaces are rotated about the $\ura{b}^1$ axis by an angle of $\pm \Gamma$. Denote this frame by $\mathcal{F}_c$. Each control surface can then rotate about the $\ura{c}^2$ axis by an angle of $\delta_\imath$, such that for each control surface, the DCM relating the segment frame to the body frame is given by $\mbf{C}_{d_\imath b} = \mbf{C}_{d_\imath c_\imath} \mbf{C}_{c_\imath b} = \mbf{C}_2(\delta_\imath) \mbf{C}_3(\Gamma_\imath)$.

The distances between the aerodynamic centres of the control surfaces and the point $z$ are set as
	\begin{align} \nonumber
		\mbf{r}_b^{c_2 z} = \begin{bmatrix} -\ell_1 & \ell_2 & -\ell_3 \end{bmatrix}^\trans, \hspace{5mm} \mbf{r}_b^{c_3 z} = \begin{bmatrix} -\ell_1 & \ell_2 & \ell_3 \end{bmatrix}^\trans, \\ \nonumber
		\mbf{r}_b^{c_4 z} = \begin{bmatrix} -\ell_1 & -\ell_2 & \ell_3 \end{bmatrix}^\trans, \hspace{5mm} \mbf{r}_b^{c_5 z} = \begin{bmatrix} -\ell_1 & -\ell_2 & -\ell_3 \end{bmatrix}^\trans.
	\end{align}

In simulation, a lateral guidance law is also used for path following. The popular guidance law presented in \cite{park2004new} is employed and is briefly explained as follows. A point on the reference path at a distance $L_1$ from point $z$ is denoted $t$. The angle between the velocity vector and line connecting $z$ to $t$ is denoted $\eta$. The guidance law generates lateral acceleration command given by
	\begin{align} \label{eq:guidance}
	\small
		a_{b,2}^{\mathrm{r}} = 2 \frac{\norm{\mbf{v}_b^{zw/a}}^2}{L_1} \sin(\eta).
	\end{align}
To achieve this lateral acceleration command, a desired roll angle can be commanded though the relation $\phi_\mathrm{r} = \frac{a_{b,2}^{\mathrm{r}}}{g}$. 
In simulation, the desired horizontal path was set as a circle centred at $\begin{bmatrix} 50 & 50 \end{bmatrix}^\trans \mathrm{m}$ with a radius of $250 \mathrm{m}$.  Though not detailed in this paper, the proportional-integral speed controller from \cite{kai2018unified} is also used to control the magnitude of the aircraft inertial velocity. Note that the control allocation problem for the thrusters is not considered in this paper and it is assumed that a collective thrust controls the magnitude of the aircraft inertial velocity. In addition, a separate control allocation problem to determine the required motor velocities to generate a desired moment while the aircraft is near hover conditions is beyond the scope of this paper. A constant desired pitch of $\theta_\mathrm{r} = 2^\circ$ is commanded. In addition, there are several methods that exist for tuning the attitude controller gains, such as classical controller design techniques based on a linearized model of the nonlinear system, as discussed in \cite{beard2012small}. However, in this case, hand-tuning was performed in simulation.

To demonstrate the effectiveness of using the wind velocity estimates directly in the attitude controller and control allocator, two simulations are performed. In the first simulation, the wind estimates are used in both the attitude controller and the control allocator. In the second simulation, no wind estimate is used. For both simulations, the initial aircraft position and velocity were set to $\mbf{r}_a^{z_0 w} = \begin{bmatrix} 0 & 0 & 0 \end{bmatrix}^\trans \mathrm{m}$ and $\mbf{v}_a^{z_0 w/a} = \begin{bmatrix} 30 & 0 & 0 \end{bmatrix}^\trans \mathrm{m}/\mathrm{s}$ respectively. The initial angular velocity and rotation vector were set to $\mbs{\omega}_b^{b_0 a} = \begin{bmatrix} 0 & 0 & 0 \end{bmatrix}^\trans \mathrm{rad}/\mathrm{s}$ and $\mbs{\phi}_0 = \begin{bmatrix} 0 & 0 & 0 \end{bmatrix}^\trans \mathrm{rad}$ respectively.  The initial wind speed is set to $\mbf{v}_a^{q_0 w/a} = \begin{bmatrix} 7 & 5 & 0.5 \end{bmatrix}^\trans \mathrm{m}/\mathrm{s}$, representing a wind speed of roughly 25 percent of the commanded UAV ground speed, which is realistic for small scale UAVs. The initial rate-gyro and accelerometer biases are set to $\mbs{\beta}_b^{1_0} = \begin{bmatrix} 0.05 & 0.1 & 0.05 \end{bmatrix}^\trans \mathrm{rad}/\mathrm{s}$ and $\mbs{\beta}_b^{2_0} = \begin{bmatrix} 0.05 & 0.05 & 0.05 \end{bmatrix}^\trans \mathrm{m}/\mathrm{s}^2$ respectively. In the initialization of the filter, the initial state estimate is randomly selected such that $\delta \mbf{X} = \mathrm{exp} \left(\delta \mbs{\xi}_0^\wedge \right)$, where $\delta \mbs{\xi}_0 \sim \mathcal{N}\left(\mbf{0}, \mbf{P}_0 \right)$, where $\mbf{P}_0$ is selected as $\mathrm{diag} \left(\mbf{1} \sigma_0^{\mbs{\xi}^\phi}, \mbf{1} \sigma_0^{\mbs{\xi}^\mathrm{v}}, \mbf{1} \sigma_0^{\mbs{\xi}^\mathrm{r}}, \mbf{1} \sigma_0^{\mbs{\xi}^\mathrm{w}}, \mbf{1} \sigma_0^{\mbs{\xi}^{\beta^1}}, \mbf{1} \sigma_0^{\mbs{\xi}^{\beta^2}}  \right)$. The values selected for the initial uncertainties are given as $\sigma_0^{\mbs{\xi}^\phi} = 10^{-2} \hspace{1mm} \mathrm{rad}$, $\sigma_0^{\mbs{\xi}^\mathrm{v}} = 10^{-2} \hspace{1mm} \mathrm{m}/\mathrm{s}$, $\sigma_0^{\mbs{\xi}^\mathrm{r}} = 10^{-5} \hspace{1mm} \mathrm{m}$, $\sigma_0^{\mbs{\xi}^\mathrm{w}} = 1 \hspace{1mm} \mathrm{m}/\mathrm{s}$, $\sigma_0^{\mbs{\xi}^{\beta^1}} = 0.05 \hspace{1mm} \mathrm{rad}/\mathrm{s}$, $\sigma_0^{\mbs{\xi}^{\beta^2}} = 0.05 \hspace{1mm} \mathrm{m}/\mathrm{s}^2$. The prediction step in the filter is performed at 1 kHz while the correction step is performed at 10 Hz. A medium-noise scenario is tested in these two simulations and the noise parameters are shown in Table~\ref{table:noise_levels} with $\gamma = 5$, where $\sigma_k^{\mathrm{Q}_\imath}, \imath = 1,2, \ldots, 5$ represents the standard deviations associated with the process model noise and $\sigma_k^{\mathrm{R}_\imath}$, $\imath = 1,2,3,4$ represents the standard deviations associated with the measurement model noise. A partial list of other simulations parameters is presented in Table~\ref{table:1}.
\vspace{-4mm}
\begin{table}[h!]
\tiny
 \caption{Noise Standard Deviations}
 \centering
    \begin{tabular}{| c | c|}
    \hline
		& Noise Intensity \\[0.8ex] \hline
		$\sigma_k^{\mathrm{Q}_1}$ (rad/s) & $ \gamma \cdot 10^{-3}$ \\[0.8ex] \hline
		$\sigma_k^{\mathrm{Q}_2}$ (m/$\mathrm{s}^2$) & $\gamma \cdot 3 \cdot 10^{-3}$ \\[0.8ex] \hline
		$\sigma_k^{\mathrm{Q}_3}$ (m/s) & $ \gamma \cdot 0.1$  \\[0.8ex] \hline
		$\sigma_k^{\mathrm{Q}_4}$ (rad/s) & $\gamma  \cdot 0.005$ \\[0.8ex] \hline
		$\sigma_k^{\mathrm{Q}_5}$ (m/$\mathrm{s}^2$) & $\gamma \cdot 0.005$  \\[0.8ex] \hline
		$\sigma_k^{\mathrm{R}_1}$ (m) & $\gamma \cdot 0.25$ \\[0.8ex] \hline
		$\sigma_k^{\mathrm{R}_2}$ (m/s) & $\gamma \cdot 0.1$  \\[0.8ex] \hline
		$\sigma_k^{\mathrm{R}_3}$ (m/s) & $\gamma \cdot 0.05$  \\[0.8ex] \hline
		$\sigma_k^{\mathrm{R}_4}$ (mG) & $\gamma \cdot 10^{-3}$ \\[0.8ex] \hline		
    \end{tabular}
    \label{table:noise_levels}
\end{table}
\vspace{-4mm}
To evaluate the performance of the attitude controller, a measure of attitude error is selected as $\Phi = \frac{1}{2} \mathrm{tr} \left(\mbf{1} - \mbfhat{C}_{ba} \mbf{C}_{ra}^\trans \right)$ \cite{lee2011geometric}. In addition, to evaluate the performance of the outer loop guidance law given by~\eqref{eq:guidance}, the cross-track error denoted $e^\mathrm{p}$ is also plotted. The cross-track error is  the lateral distance between the aircraft and the closest point on the path at a given time, resolved in the Frenet-Serret frame defined by the path at that point\cite{kai2018unified}.

\begin{figure}[H]
			\centering
        \includegraphics[width=0.38\textwidth]{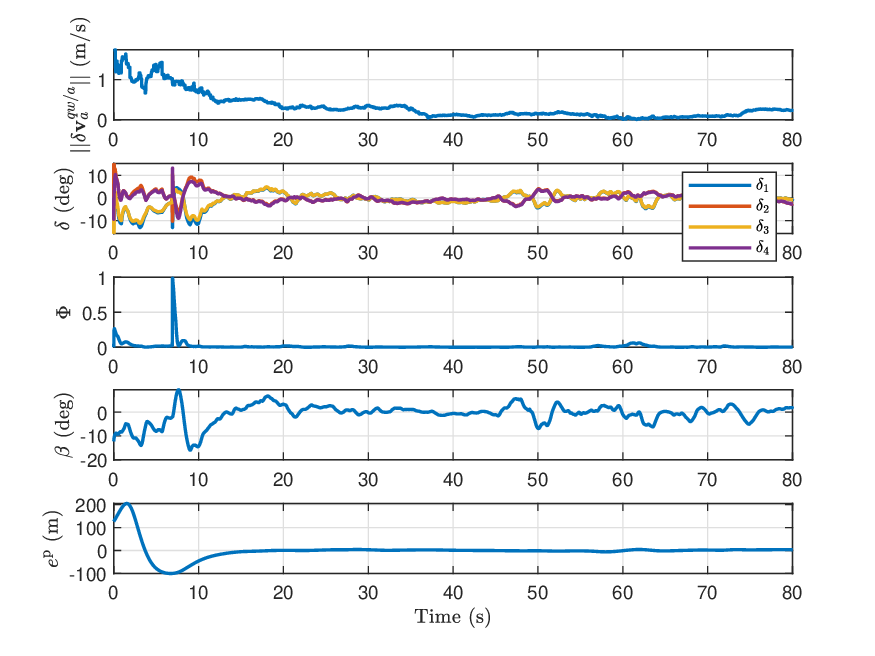}
        \vspace{-7mm}
		\caption{Simulation 1, wind estimates used in controller.}
		\label{fig:wind_estimate}
		\end{figure}

\begin{figure}[H]
			\centering
        \includegraphics[width=0.42\textwidth]{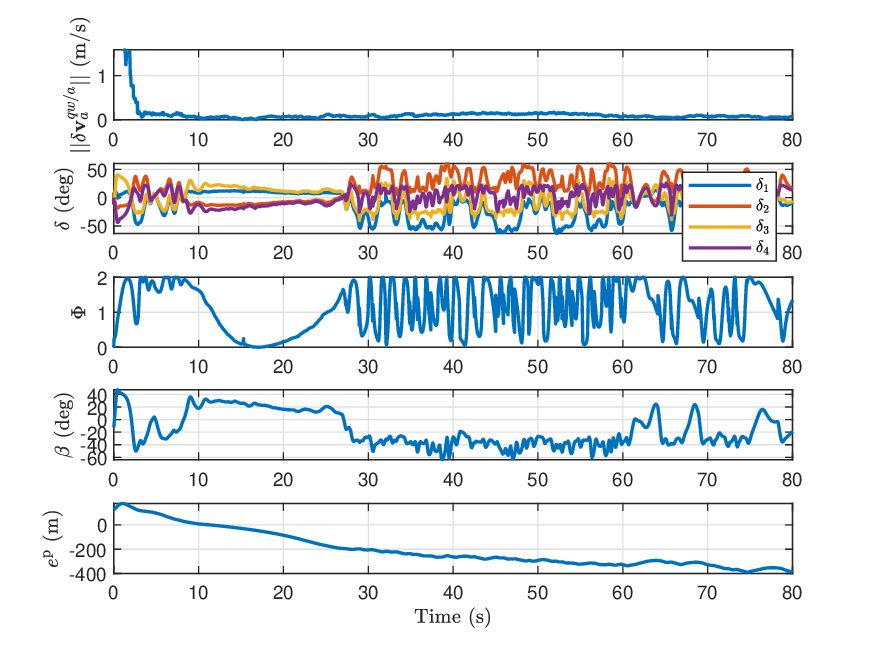}
        \vspace{-5mm}
		\caption{Simulation 2, wind estimates not used in controller.}
		\label{fig:bad_results}
		\end{figure}

Figure~\ref{fig:wind_estimate} shows the error in estimated wind velocity, the control surface deflections, the attitude error $\Phi$, the sideslip angle, and the cross-track error when wind estimates are used in the controller and control allocator, and Figure~\ref{fig:bad_results} shows the same variables when no wind estimate is used.
\vspace{-4mm}
		\vspace{-5mm}
		\vspace{-8mm}
The spike in the attitude error function $\Phi$ in Figure~\ref{fig:wind_estimate} at approximately 8 seconds is due to a rapid change in the desired lateral acceleration command output by the guidance law, leading to a change in the reference roll angle and the reference DCM $\mbf{C}_{ra}$.
		
It is seen that when the wind velocity estimate is not used in the controller, the attitude cannot be regulated to the desired attitude, and thus the sideslip angle is not regulated and performance in the outer loop degrades. When the estimated wind velocity is used in the controller, the attitude error and sideslip are both regulated, and the UAV converges to the circular path.

\begin{figure}[H]
			\centering
        \includegraphics[width=0.42\textwidth]{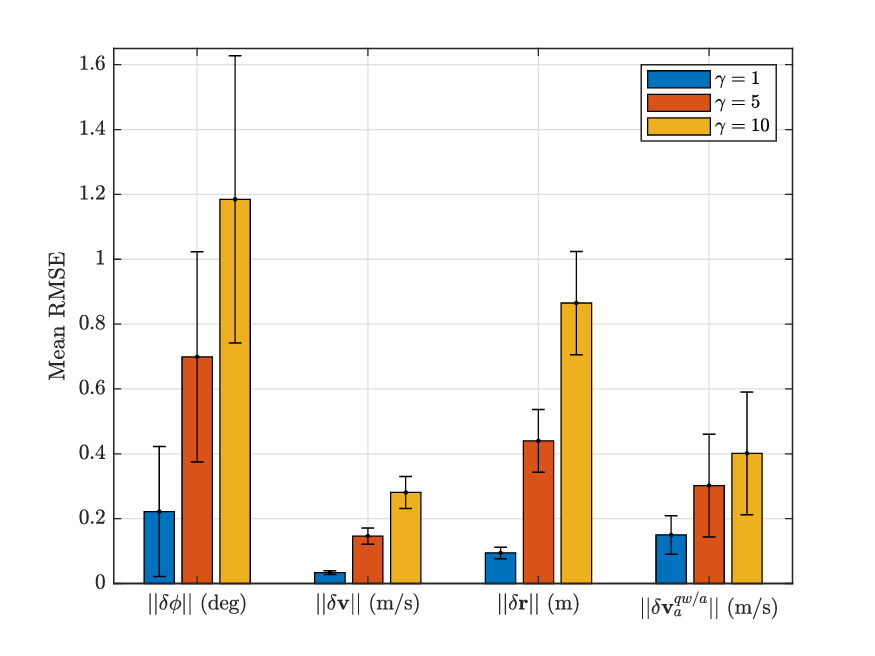}
        \vspace{-5mm}
		\caption{Monte-Carlo results for the estimation error.}
		\label{fig:monte_carlo_estimation}
		\end{figure}

To test the robustness of the controller and estimator to various filter initializations, noise intensity levels, and state trajectories, Monte-Carlo simulations were performed with the wind velocity estimate being used directly in the controller and control allocator. To test the robustness of the filter to initialization error, at each trial, the initial estimate was chosen randomly such that $\delta \mbf{X}_0 = \mathrm{exp} \left(\delta \mbs{\xi}_0^\wedge \right)$, where $\delta \mbs{\xi}_0 \sim \mathcal{N}\left(\mbf{0}, \mbf{P}_0 \right)$, where $\mbf{P}_0$ is selected as in the previous simulations. In addition, to test the robustness of the filter to various state trajectories, as well as the robustness of the control scheme, the initial attitude of the aircraft is set to $\mbf{C}_{b_0 a} = \mbf{C}_3(\psi_0)$, where $\psi_0$ is a random initial heading angle of the aircraft with $\psi_0 \sim \mathcal{N}\left(0, \pi \right)$ for each trial. The initial velocity for each trial is then set to $\mbf{v}_a^{z_0 w/a} = \mbf{C}_{b_0 a}^\trans \begin{bmatrix} 30 & 0 & 0 \end{bmatrix}^\trans \mathrm{m}/\mathrm{s}$. Three noise levels were tested corresponding to low, medium, and high sensor noise levels and noise levels on the wind and bias random walks. The standard deviations on the process and measurement model noise for each of the three noise trials are shown in Table~\ref{table:noise_levels}. 50 Monte-Carlo simulations were performed at each $\gamma = 1, 5, 10$, to simulate low, medium and high sensor noise levels, respectively. For each noise level, the mean RMSE of the estimation errors $\norm{\delta \mbs{\phi}}$, $\norm{\delta \mbf{v}}$, $\norm{\delta \mbf{r}}$, and $\norm{\delta \mbf{v}_a^{qw/a}}$ are taken over all Monte-Carlo trials and the results as well as the $2 \sigma$ bounds are plotted in Figure~\ref{fig:monte_carlo_estimation}. To ensure that 50 Monte-Carlo simulations at each noise level accurately captures the performance of the system, the statistics of 50 trials were compared to the statistics of 40 trials, and marginal difference between mean RMSEs of the errors and standard deviations were observed. 

\begin{figure}[H]
			\centering
        \includegraphics[width=0.42\textwidth]{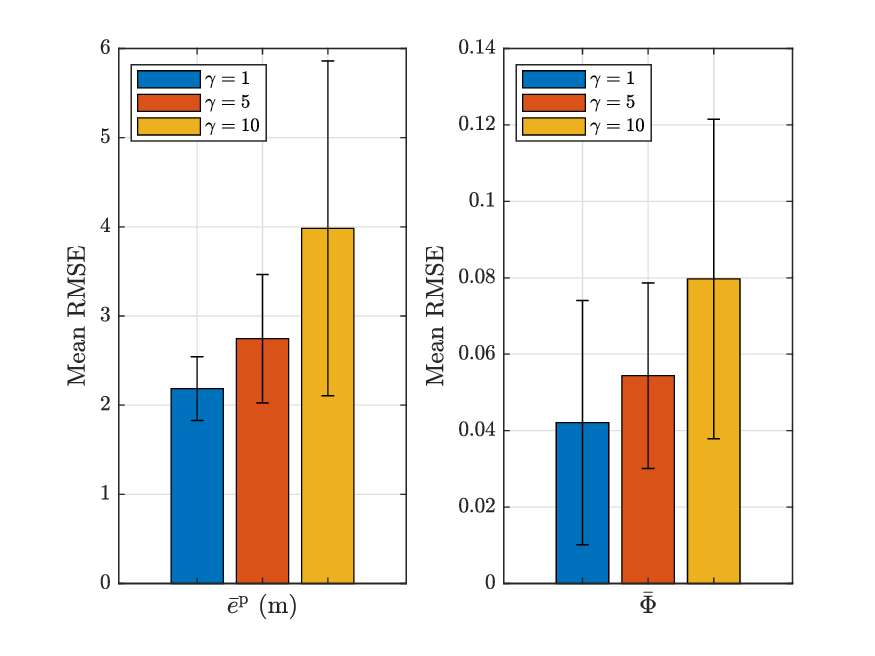}
        \vspace{-5mm}
		\caption{Monte-Carlo results for position and attitude tracking errors.}
		\label{fig:monte_carlo_control}
		\end{figure}

\begin{figure}[H]
			\centering
        \includegraphics[width=0.35\textwidth]{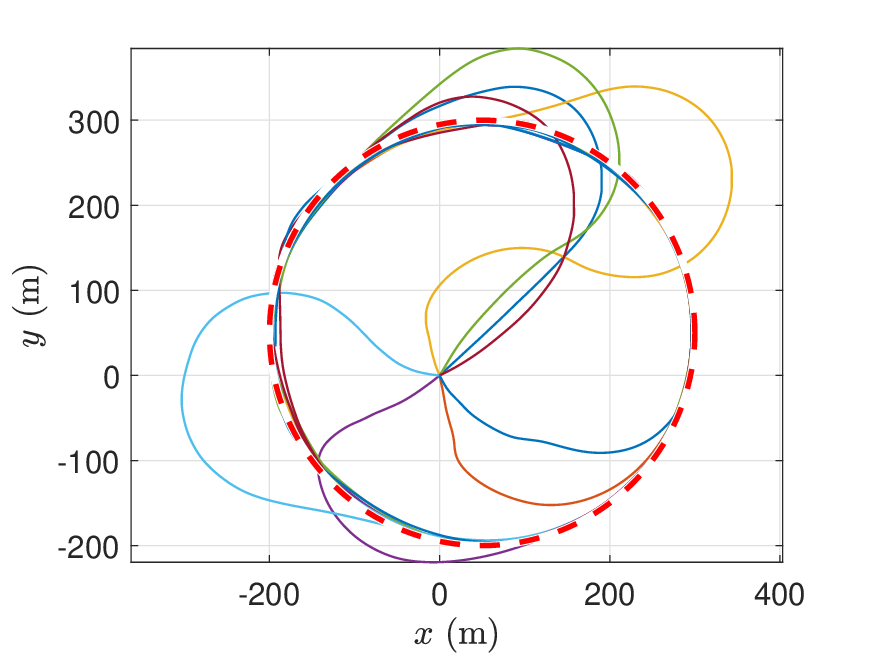}
        \vspace{-4mm}
		\caption{UAV position in the $x$-$y$ plane for seven sample Monte-Carlo simulations, $\gamma = 5$.}
		\label{fig:monte_carlo_position}
		\end{figure}

		\vspace{-5mm}
To evaluate the performance of the guidance law and attitude controller for various levels of process and measurement model noise, filter initializations, and initial heading angles, the mean RMSE of the cross-track error after convergence to the path and the mean RMSE of the attitude error function as well as the $2 \sigma$ bounds are shown in Figure~\ref{fig:monte_carlo_control}. Note that convergence to the path has been defined as the point at which the cross-track error is below $10 \mathrm{m}$ and never becomes higher than $10 \mathrm{m}$ again. 	
	\vspace{-6mm}
The position of the UAV in the $x$-$y$ plane for seven runs of the Monte-Carlo simulations with $\gamma = 5$ is shown in Figure~\ref{fig:monte_carlo_position}. The dotted red line represents the desired circular path to be followed by the UAV and each colored line represents the path taken in one simulation. Note that all seven runs converge to the desired path.
\vspace{-2mm}
\vspace{-5mm}
When the wind velocity estimate is not used in the control, the UAV does not converge to the path regardless of the filter initialization error, noise intensities and state trajectories. Thus, Monte-Carlo analysis is not shown for the case that the wind estimate is not used directly in the control.
	
Several other aircraft configurations were tested in simulation, and it was observed that some configurations of control surfaces lead to a control allocation scheme that is much less robust to error in state and wind estimates than others. Thus, the presented control and estimation strategies can be used to ensure that the proposed geometries of VTOL UAVs lead to control that is still robust to error in state estimates.
\begin{table}[h!]
\tiny
 \caption{Parameter List}
 \centering
    \begin{tabular}{ |l | l |l  |}
    \hline
     Parameter & Value & Units\\ \hline
    $m_\mathcal{B}$ & 8.26 & \si{\kilogram} \\ \hline
    $\mbf{J}_b^{\mathcal{B} z} $ & diag(1.42, 0.82, 1.75) & \si{\kilogram \meter\squared} \\ \hline
    $L_1$ & 120 & \si{\meter} \\ \hline   
    $\ell_1, \ell_2, \ell_3$ & 0.5, 0.25, 0.4 & \si{\meter} \\ \hline 
    $\mbf{K}^{\mbs{\phi}}$ & diag(100, 100, 100) & \si{\newton\per\meter} \\ \hline 
    $\mbf{K}^{\mbs{\omega}}$ & diag(50, 50, 50) & \si{\newton\meter\second} \\ \hline 
    $\mbf{K}^{i,1}$ & diag(10, 10, 15) & \si{\newton\meter\per\second} \\ \hline 
    $\mbf{K}^{i,2}$ & diag(1, 2, 5) & \si{\newton\per\meter} \\ \hline 
    $\Gamma$ & 35 & deg \\ \hline
    \end{tabular}
    \label{table:1}
\end{table}
\normalsize

\section{Conclusions and Future Work} \label{sec:conclusion}
The estimation and control problems for a class of unconventional VTOL UAVs operating in forward-flight conditions were presented in this paper. A tightly-coupled estimation approach is employed to estimate the aircraft navigation states as well as the wind velocity. The estimation solution is done in an IEKF framework, which offers advantages compared to the more common MEKF. The wind velocity estimate is then used to improve the performance of an $SO(3)$-based attitude controller and a control allocator. There are various avenues to explore in the future. For example, given that a mix of sensors are used, comparing the performance of the left-invariant ``Imperfect" IEKF to a right-invariant one, as well as to a left- and right-invariant sigma point Kalman filter, would be interesting. To more accurately account for nonlinear post-stall aerodynamics and prop-wash, higher-fidelity modeling may be of interest. The development of guidance strategies that directly use the wind velocity estimates can also be investigated.

\appendices


\section*{Acknowledgment}

The authors would like to thank Guillaume Charland-Arcand for his help and support.

\ifCLASSOPTIONcaptionsoff
  \newpage
\fi



%

\addcontentsline{toc}{section}{References}
\bibliographystyle{ieeetr}

\bibliography{ref}

%




\end{document}

%% file: cover.tex
%
%
%
%
%
%
%
\def \myJournal {IEEE Robotics and Automation Letters}
\def \myDoi {10.1109/LRA.2020.2966406}
\def \myPaperSiteName {IEEE Xplore}
\def \myPaperSiteLink {https://ieeexplore.ieee.org/document/8957471}
\def \myYear {2024}

\def \myPaperCitation{M. Cohen and J. R. Forbes, ``Navigation and Control of
Unconventional VTOL UAVs in Forward-Flight with Explicit Wind
Velocity Estimation,'' in \textit{IEEE Robotics and Automation Letters}, vol. 5, no. 2, pp. 1151-1158, January 2020.}


\begin{figure*}[t]

\thispagestyle{empty}
\begin{center}
\begin{minipage}{6in}
\centering
This paper has been accepted for publication in \emph{\myJournal}. 
\vspace{1em}

This is the author's version of an article that has, or will be, published in this journal or conference. Changes were, or will be, made to this version by the publisher prior to publication.
\vspace{2em}

\begin{tabular}{rl}
DOI: & \myDoi\\
\myPaperSiteName: & \texttt{\myPaperSiteLink}
\end{tabular}

\vspace{2em}
Please cite this paper as:

\myPaperCitation

\vspace{15cm}
\copyright \myYear \hspace{4pt}IEEE. Personal use of this material is permitted. Permission from IEEE must be obtained for all other uses, in any current or future media, including reprinting/republishing this material for advertising or promotional purposes, creating new collective works, for resale or redistribution to servers or lists, or reuse of any copyrighted component of this work in other works.

\end{minipage}
\end{center}
\end{figure*}
\newpage
\clearpage
\pagenumbering{arabic} 